\pdfoutput=1

\documentclass[11pt]{article}

\usepackage[final]{coling}

\usepackage{times}
\usepackage{latexsym}

\usepackage[T1]{fontenc}

\usepackage[utf8]{inputenc}

\usepackage{microtype}

\usepackage{inconsolata}

\usepackage{times}
\usepackage{latexsym}
\usepackage{algorithm}
\usepackage{algorithmic}
\usepackage{graphicx}
\usepackage{booktabs}
\usepackage{algorithm}
\usepackage{algorithmic}
\usepackage{amsfonts,amssymb}
\usepackage{amsmath,bm}
\usepackage{subcaption}
\usepackage{diagbox}
\usepackage{url}
\usepackage{multirow} 
\usepackage{multicol} 
\usepackage{xcolor}
\usepackage{colortbl}
\usepackage{mathrsfs}
\usepackage{tabularx}
\usepackage{makecell}
\usepackage{amsfonts}
\usepackage{bbm}
\usepackage{arydshln}
\usepackage{booktabs}
\usepackage{bbding}
\usepackage{pifont}

\usepackage{listings}
\newcommand{\datasetname}{MIT-10M}
\newcommand{\llavanext}{LLaVA-NeXT}
\newcommand{\minicpm}{MiniCPM-Llama3-V}
\newcommand{\deepseekvl}{DeepSeek-VL}
\newcommand{\internvl}{InternVL2}
\newcommand{\cogvlm}{CogVLM2-LLaMA3}
\newcommand{\qwenvl}{Qwen2-VL}

\title{\datasetname: A Large Scale Parallel Corpus of  Multilingual Image Translation}
\author{Bo Li $^{1,3}$, Shaolin Zhu$^{2}$ \footnotemark[1], Lijie Wen$^1$ \thanks{The corresponding author.}\\
$^1$School of Software, Tsinghua University, Beijing, China  \\
$^2$College of Intelligence and Computing, Tianjin University, Tianjin, China\\
$^3$Baidu Inc., Beijing, China\\
\texttt{libo15@baidu.com, zhushaolin@tju.edu.cn, wenlj@tsinghua.edu.cn}
}

\begin{document}
\maketitle
\begin{abstract}

Image Translation (IT) holds immense potential across diverse domains, enabling the translation of textual content within images into various languages.
However, existing datasets often suffer from limitations in scale, diversity, and quality, hindering the development and evaluation of IT models.
To address this issue, we introduce \datasetname~\footnote{Our datasets and code are publicly available at: \url{https://huggingface.co/datasets/liboaccn/MIT-10M}}, a large-scale parallel corpus of multilingual image translation with over 10M image-text pairs derived from real-world data, which has undergone extensive data cleaning and multilingual translation validation. 
It contains 840K images in three sizes, 28 categories, tasks with three levels of difficulty and 14 languages image-text pairs, which is a considerable improvement on existing datasets.
We conduct extensive experiments to evaluate and train models on \datasetname. 
The experimental results clearly indicate that our dataset has higher adaptability when it comes to evaluating the performance of the models in tackling challenging and complex image translation tasks in the real world. 
Moreover, the performance of the model fine-tuned with \datasetname~has tripled compared to the baseline model, further confirming its superiority.
\end{abstract}

\begin{figure}[t]
    \centering
    \includegraphics[width=0.85\columnwidth]{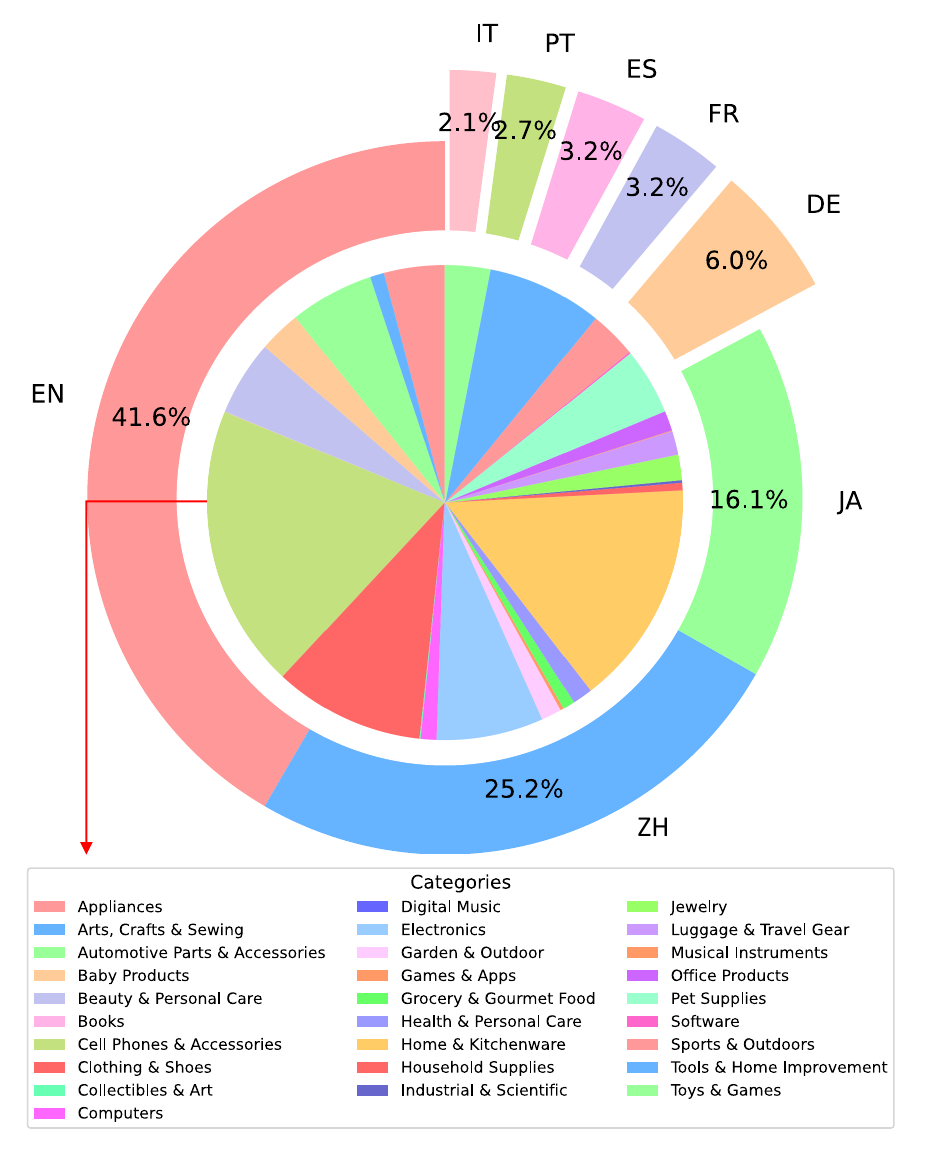}
    \caption{Categories and languages of \datasetname. It includes 28 categories and 14 languages image-text pairs (8 languages images).}
    \label{fig:diversity}
\end{figure}

\section{Introduction}

\begin{figure*}[t]
\centering
    \includegraphics[width=0.95\textwidth]{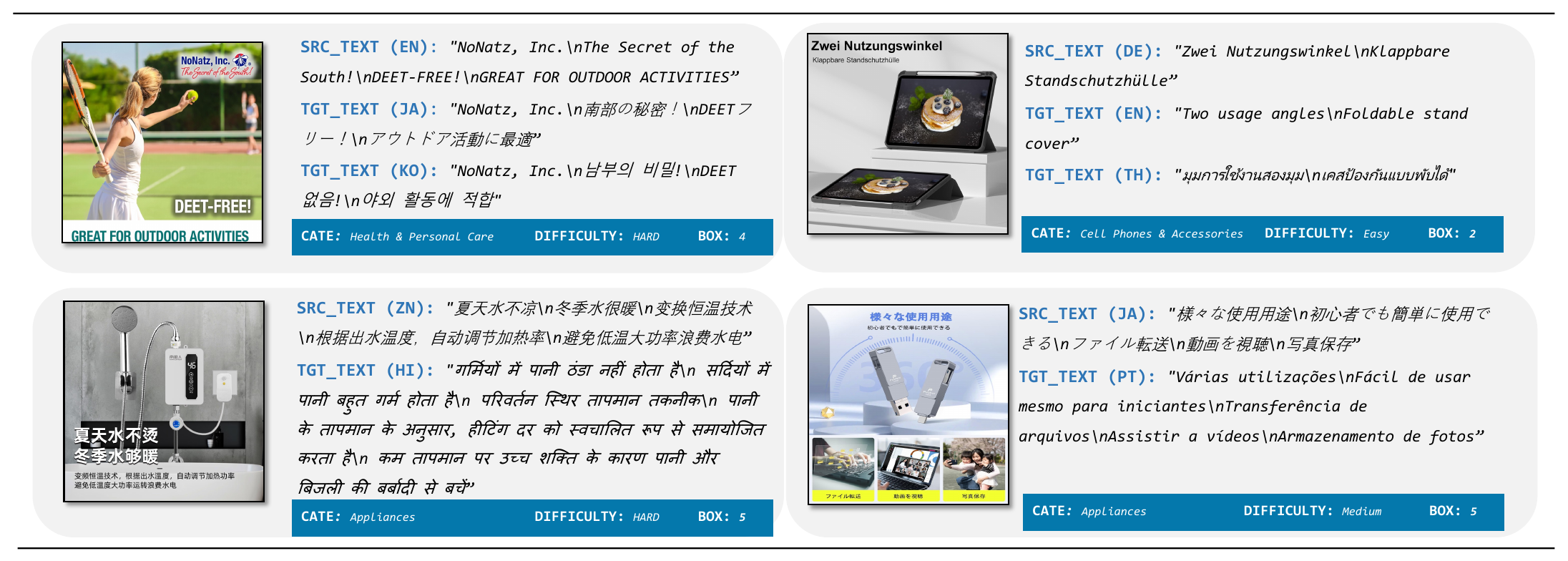}
\caption{Examples of \datasetname~dataset. Each image contains the original text and the corresponding language. Additionally, we annotate the image category, the token length of the text and the number of bounding boxes and used them for the difficulty level. In addition to the original text, 13 languages translations were annotated.}
\label{fig:img_demo}
\end{figure*}

Image Translation (IT), the task of translating embedded text within an image from a source language to a target language \cite{watanabe1998translation,yang2002automatic,lan2023exploring}, holds significant promise for various applications. 
Its utility spans domains such as scene text translation, document image translation \cite{liang-etal-2024-document}, and photo translation, enhancing accessibility and cross-lingual communication. 
This is in contrast to traditional neural machine translation \cite{chen-etal-2022-synchronous,zhu-etal-2024-landermt}, which is based solely on textual information.

Traditional IT approaches often employed cascade methods \cite{zhang2019sequence,zhao-etal-2020-knowledge,shekar2021optical,hinami2021towards,afli-way-2016-integrating}.
The emergence of end-to-end IT models \cite{zhu-etal-2023-peit,liang-etal-2024-document,niu-etal-2024-umtit,ma2023e2timt} offered a more streamlined approach, utilizing a unified model to directly translate image text.  
Multimodal Large Language Models (MLLMs)  \cite{bai2023qwen,chen2024internvl,hong2024cogvlm2,liu2024llavanext,yao2024minicpm,lu2024deepseek,li2023blip} have further fueled progress.

However, IT model development and evaluation lies in the scarcity of high-quality, large-scale datasets  \cite{Ma2022ImprovingET,ma2023multi,zhu-etal-2023-peit,ma2023e2timt,liang-etal-2024-document}.  
Models trained on limited real-world data often struggle to generalize to complex scenarios, while those trained on synthetic datasets  \cite{9412281,su2021rtnet,niu-etal-2024-umtit} may not adequately capture the nuances of real-world image characteristics. 
Furthermore, existing datasets frequently lack fine-grained splits and diversity in language representation and task difficulty, hindering comprehensive model assessment.

To address these issues, we introduce \datasetname, a large-scale multilingual image translation corpus.
\datasetname~is the largest real-world image translation dataset to date.
As shown in Figure \ref{fig:diversity}, it includes 840K images in 8 languages and 10M image-text pairs across 14 languages, and split into a train set a test set.
Figure \ref{fig:img_demo} shows some examples from the \datasetname~dataset. 
Each image contains the original text and the corresponding language. 
Additionally, we annotate the image category, the token length of the text and the number of bounding boxes and used them for the difficulty level. 
In addition to the original text, 13 further translations were annotated. Detailed examples can be found in Appendix \ref{sec:appendix_examples}.
The \datasetname~dataset construction process consists of three main stages: data collection and pre-processing, OCR annotation and cleaning, and multilingual translation and validation.
We compared \datasetname~with several existing popular image  translation datasets in terms of data scale, task difficulty level, diversity, and image quality. 
The comparison results demonstrate that \datasetname~significantly outperforms other datasets in these aspects.
We conduct extensive experiments to evaluate the multilingual translation capabilities of 7 end-to-end IT models using the \datasetname~test set. 
The results show that using the \datasetname~test set is beneficial when evaluating challenging tasks of translating multi-line text into images with complex backgrounds.
Furthermore, we fine-tune the \qwenvl~\cite{bai2023qwen} model using the \datasetname~training set, BLEU, chrF++, and METEOR scores have increased by 230\%, 88\%, and 130\% respectively.
The results demonstrate significant performance improvements in multilingual image translation tasks, further validating the advantages of \datasetname~dataset.

The main contributions are outlined below:
\begin{itemize}
    \item We present \datasetname, a dataset comprising 10M image-text pairs and 840K high-resolution real-world images, representing the largest publicly available real-world high-quality dataset specifically designed for multilingual image translation tasks.
    
    \item We propose a multi-dimensional evaluation framework for multilingual image-text translation datasets, considering aspects such as data scale, task difficulty level, diversity, and image quality. Our comparative analysis demonstrates the superiority of \datasetname~across these dimensions.
    
    \item We conduct comprehensive experiments to validate the effectiveness of \datasetname~for training and evaluating multilingual image translation models.
\end{itemize}

\begin{figure*}[t]
\centering
    \includegraphics[width=0.95\textwidth]{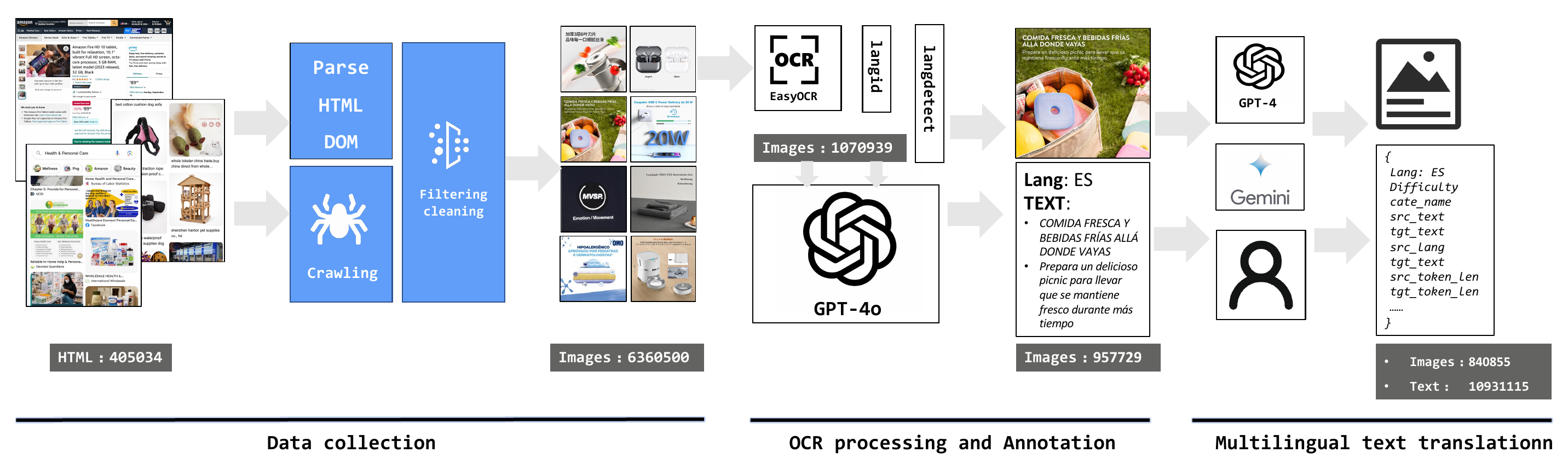}
\caption{Overview of \datasetname~dataset construction pipeline.}
\label{fig:img_arch}
\end{figure*}

\section{Related Work}
IT focuses on translating text embedded within images from a source language into a target language.  
This section reviews existing approaches and datasets relevant to IT.

\subsection{IT Models}

Two primary paradigms dominate the field of image translation:
\paragraph{Cascade Methods.} These approaches decompose the task into sequential steps, typically employing Optical Character Recognition (OCR) followed by NMT \cite{zhao-etal-2020-knowledge,shekar2021optical,hinami2021towards,zhong-etal-2024-context,chen-etal-2024-dual,zhu-etal-2024-landermt}. 
While conceptually straightforward, cascade methods suffer from error propagation between stages, increased latency due to separate model processing, and redundant parameterization \cite{ma2023multi,Ma2022ImprovingET,lan2023exploring,zhu-etal-2023-peit}.
\paragraph{End-to-End Methods.} These models aim to directly translate image text using a unified architecture \cite{mansimov-etal-2020-towards,jain2021image,zhu-etal-2023-peit,liang-etal-2024-document,niu-etal-2024-umtit,ma2023e2timt}, typically comprising a visual encoder for extracting image features and a text decoder for generating the target translation. 
The advent of Multimodal Large Language Models (MLLMs), such as  \cite{bai2023qwen,chen2024internvl,hong2024cogvlm2,liu2024llavanext,yao2024minicpm,lu2024deepseek,li2023blip}, has significantly advanced end-to-end IT, enabling more effective cross-modal fusion and improved translation accuracy. 
However, the success of end-to-end models heavily relies on the availability of large-scale, high-quality parallel image-translation data, which remains a critical challenge in the field \cite{Ma2022ImprovingET,ma2023multi,zhu-etal-2023-peit,ma2023e2timt,liang-etal-2024-document}.

\subsection{IT Datasets}

Due to the scarcity of real-world image translation data, several studies have utilized synthetic datasets created by rendering text onto background images. 
These include datasets focused on single-line text \cite{mansimov-etal-2020-towards}, multi-line text \cite{jain2021image}, Chinese-English translation \cite{Ma2022ImprovingET}, and English-German translation \cite{niu-etal-2024-umtit}. While valuable for initial model training, synthetic datasets often oversimplify real-world complexities and may not adequately reflect the diversity and challenges encountered in practical applications.
Manually curated real-world datasets are crucial for advancing IT research, but their development is resource-intensive, resulting in limited availability.  
Existing datasets include OCRMT30K \cite{lan-etal-2024-translatotron}, a Chinese-English dataset based on OCR annotations;  DoTA \cite{liang-etal-2024-document}, focused on document image translation into Markdown format; and ECOIT \cite{zhu-etal-2023-peit}, targeting the e-commerce domain.  
Despite these efforts, real-world datasets remain limited in scale, language coverage, and diversity of image characteristics, hindering comprehensive model evaluation and generalization assessment. 
Our work addresses this gap by introducing \textbf{\datasetname}, a novel large-scale, multilingual, and real-world image translation dataset designed to facilitate the development and evaluation of robust and generalizable IT models.

\begin{table*}[t]
\centering
\renewcommand{\arraystretch}{1.1}
\small
\resizebox{0.9\textwidth}{!}{
\begin{tabular}{lccrr}
\toprule
\textbf{Dataset} & \textbf{Source} & \textbf{Languages} & \textbf{Images} & \textbf{Image-Text} \\ \hline
E2E\_TIT\_With\_MT \cite{Ma2022ImprovingET}  & synthetic  & 3  & 3,000,000  & 3,000,000  \\  
ECOIT  \cite{zhu-etal-2023-peit}    & realistic  & 2  & 479,490    & 479,490    \\  
OCRMT30K   \cite{lan2023exploring} & realistic  & 2  & 30,000    & 30,000    \\  
DoTA  \cite{liang-etal-2024-document}    & realistic  & 2  & 126,345    & 126,000    \\  
IIMT  \cite{lan-etal-2024-translatotron}   & synthetic  & 1  & 89,033     & 89,033     \\ \hline
\cellcolor[gray]{0.92}\textbf{\datasetname~(Ours)}      & \cellcolor[gray]{0.92}\textbf{realistic} & \cellcolor[gray]{0.92}\textbf{14} & \cellcolor[gray]{0.92}\textbf{840,855} & \cellcolor[gray]{0.92}\textbf{10,931,115} \\ \bottomrule
\end{tabular}
}
\caption{Comparison of \datasetname~with other popular image translation datasets in terms of statistical data.}
\label{tab:dataset_comparison}
\end{table*}

\section{\datasetname~Construction}

In this section, the construction pipeline of \datasetname~is described in detail, highlighting the crucial filtering steps to ensure data quality.
Figure \ref{fig:img_arch} illustrates the \datasetname~construction pipeline.

\subsection{Data collection}
\paragraph{Web Crawling.}

High-quality websites with extensive language coverage were selected for data collection, including google.com, baidu.com, amazon.com, jd.com, and amazon.jp.co. These websites offer a rich source of high-resolution images and support multiple languages.  
Crawling was conducted across eight languages: English, French, Chinese, Japanese, Portuguese, Italian, German, and Spanish. 
To ensure data uniqueness, SHA256 hashing was employed to identify and remove duplicate pages. 
Documents lacking images or containing an excessive number of images (over 50) were excluded. 
This process resulted in 405K unique HTML files, occupying 440 GB.

\paragraph{Image collection.}
We use the BeautifulSoup \cite{bs} library to parse the HTML documents into a tree structure and extract the image tags.
To ensure the quality of the data, we filter the images according to their resolution and only keep the images with a resolution of more than 800x800 pixels that are in the main content area of the HTML document to exclude logos and advertisements.
Duplicate images are then removed based on their MD5 hash values.

\paragraph{Filtering and cleaning data.}
To ensure that the content is appropriate, we apply a NSFW detection tool \cite{man} to all images.
If NSFW content is detected, we discard all images from the corresponding HTML document.
The final set contains 6.3M images occupying 900 GB.

\subsection{Data annotation}
\paragraph{Initial text recognition.}
To quickly determine whether images contain text, we first apply easyOCR\footnote{https://github.com/JaidedAI/EasyOCR} for text recognition and remove images where no text is recognized or the recognized text is meaningless, which accounts for about 50\% of the images.
This results in 3M images being retained.
Next, we use the tools langid\footnote{https://github.com/saffsd/langid.py} and langdetect\footnote{https://github.com/fedelopez77/langdetect} to identify the language of the recognized text and perform cross-validation.
Only images for which both tools consistently identify the same language are retained.
The result is 1 million images.

\paragraph{Precise text recognition.}
We then use GPT-4o \cite{gpt-4o} for detailed text recognition to extract precise text information from the images and discard those images where the text cannot be recognized.
\paragraph{Filtering sensitive content and text cleaning.}
To reduce the risk of sharing personal data, we remove images whose OCR-recognized text contains sensitive information such as email addresses or IP addresses.
We also exclude images where the recognized text contains NSFW characters to exclude potentially inappropriate content.
To further improve the quality of the dataset, we filter out images with advertising or meaningless content (e.g. excessive numbers or punctuation) and remove images with excessively long text (over 450 tokens or 60 words).
After the cleanup, 957K images remain, taking up 150 GB.

\subsection{Multilingual text translationn}
\paragraph{Machine translation.}
To translate the text into other languages, we first use GPT-4 \cite{achiam2023gpt} to translate the OCR-recognized text.
We create a highly optimized translation prompt that translates the image text from 8 languages into 13 languages, with the additional languages being Korean, Thai, Arabic, Turkish, Hindi and Russian.
\paragraph{Validation of the translation.}
Although GPT-4 performs well in multilingual translation, we cross-validate with Google Translate (via the Google Gemini 1.5 Pro API) \cite{geminiteam2024gemini15unlockingmultimodal}.
We use the tool spacy to convert the results of GPT-4 and Google Translate into word vectors and calculate their semantic similarity.
For text pairs with a similarity score below 0.8, we filter out those with significant differences and keep translations with a semantic similarity above 0.8.
The final parallel corpus contains 840K images.
\paragraph{Human evaluation.}
We divide the translated image-text pairs and their corresponding text translations into 10,000 batches. We randomly select 10 batches and perform a human evaluation, which results in a translation accuracy of 99.4\%.

\begin{figure}[t]
    \centering
    \includegraphics[width=1\columnwidth]{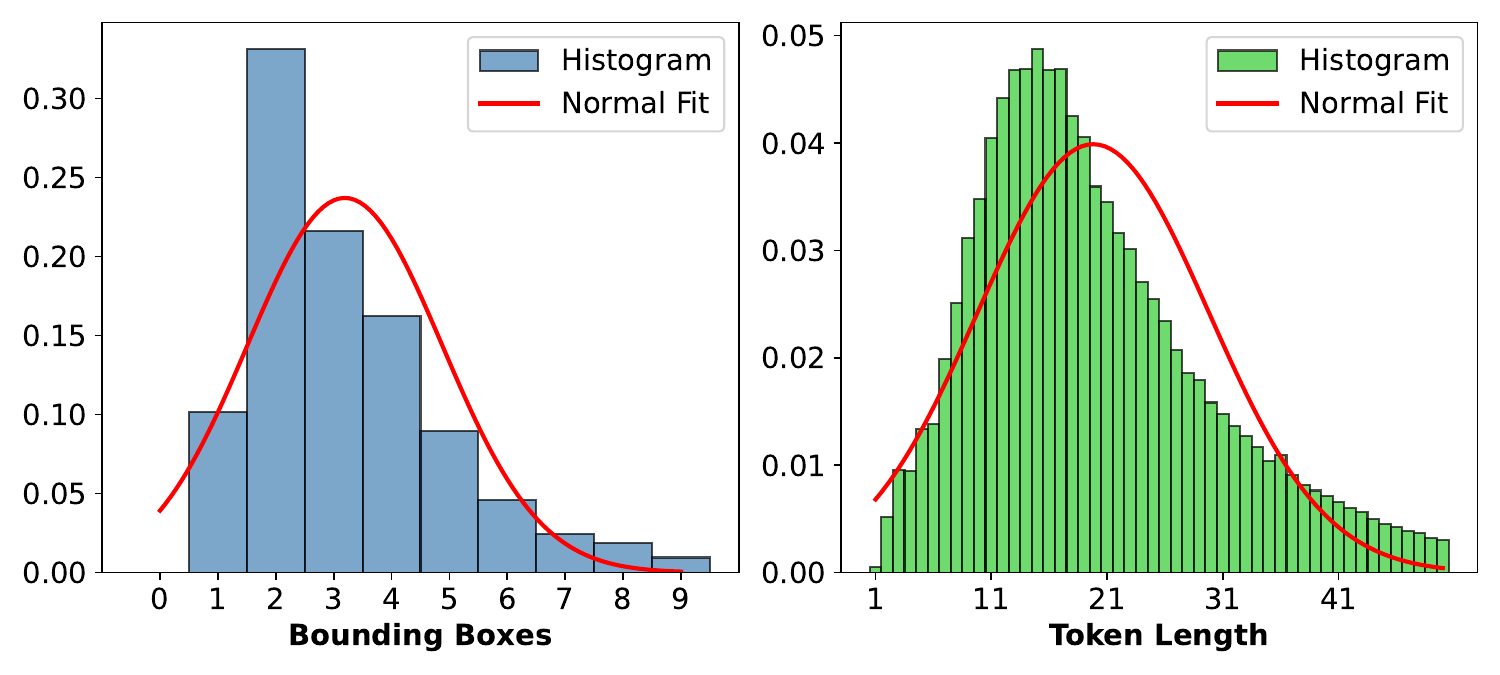}
    \caption{Distribution of the number of bounding boxes in the images and the token lengths in English text. }
    \label{fig:block_and_token_len}
\end{figure}

\section{Analysis \datasetname}

This section presents a comprehensive analysis of the \datasetname~dataset. First, we compare \datasetname~with existing popular image translation datasets (see Table \ref{tab:dataset_comparison}). 
Then, we analyze \datasetname~in detail in terms of data scale, difficulty, diversity, and image quality.

\subsection{Data Scale Comparison}
Table \ref{tab:dataset_comparison} shows a comparison of \datasetname~with other popular image translation datasets in terms of statistical data. \datasetname~clearly outperforms the other datasets in terms of the number of languages, images and image-text pairs.

\datasetname~includes 14 languages and is therefore better suited for cross-language multimodal image translation research, especially for scenarios requiring the processing of multiple languages. In addition, \datasetname~is derived from real-world scenarios and contains 840K images and 10M image-text pairs. This is a significantly larger volume than data sets such as DoTA and ECOIT, which also consist of real images.

\subsection{Difficulty Levels}
\label{sec:diff_levels}

\begin{table}[t]
    \centering
    \renewcommand{\arraystretch}{1.1}
    \small
    \resizebox{0.95\columnwidth}{!}{
    \begin{tabular}{lcc}
       \toprule
        & \textbf{@Train} & \textbf{@Test} \\ \hline
        \textbf{Easy}   & 3,154,034 (28.9\%) & 3,120 (30.0\%) \\ 
        \textbf{Medium} & 3,156,621 (28.9\%) & 3,120 (30.0\%) \\ 
        \textbf{Hard}   & 4,610,060 (42.2\%) & 4,160 (40.0\%) \\ 
        \bottomrule
    \end{tabular}
    }
    \caption{The number and proportion of corpora of different difficulty levels of \datasetname.}
    \label{tab:data_distribution}
\end{table}

\begin{table}[ht]
\centering
\resizebox{0.9\linewidth}{!}{
\begin{tabular}{cll}
\toprule
\textbf{Cate ID} & \textbf{Category Name} & \textbf{Images} \\
\midrule
10000  & Appliances                    & 34711  \\
11000  & Digital Music                 & 44     \\
12000  & Electronics                   & 60884  \\
13000  & Garden \& Outdoor              & 11559  \\
14000  & Games \& Apps                 & 1940   \\
15000  & Grocery \& Gourmet Food        & 6654   \\
16000  & Health \& Personal Care        & 11447  \\
17000  & Home \& Kitchenware            & 129192 \\
18000  & Household Supplies             & 4298   \\
19000  & Industrial \& Scientific       & 1486   \\
20000  & Arts, Crafts \& Sewing         & 8077   \\
21000  & Luggage \& Travel Gear         & 13643  \\
22000  & Musical Instruments            & 451    \\
23000  & Office Products               & 11462  \\
24000  & Pet Supplies                  & 38158  \\
25000  & Software                      & 977    \\
26000  & Sports \& Outdoors             & 26742  \\
27000  & Tools \& Home Improvement      & 65771  \\
28000  & Toys \& Games                 & 25776  \\
29000  & Computers                     & 9039   \\
30000  & Automotive Parts \& Accessories & 47869  \\
31000  & Jewelry                       & 14538  \\
40000  & Baby Products                 & 23931  \\
50000  & Beauty \& Personal Care        & 42964  \\
60000  & Books                         & 211    \\
70000  & Cell Phones \& Accessories     & 162428 \\
80000  & Clothing \& Shoes              & 85998  \\
90000  & Collectibles \& Art            & 605    \\
\bottomrule
\end{tabular}
}
\caption{Distribution of image category.}
\label{tbl:appendix_data_cate}
\end{table}

We visualized the distribution of the number of bounding boxes in the images and the corresponding lengths of the English tokens in the \datasetname~dataset, as shown in Figure~\ref{fig:block_and_token_len}. 
Most of the images have a bounding box count between 1 and 3, with the proportion of images with a bounding box count of 2 being the highest. In addition, the length of text tokens in the images is mainly in the range of 10 to 25 tokens, which is consistent with the distributional characteristics of text length in natural language.

Based on the above analysis, we categorize the \datasetname~dataset into three difficulty levels: easy, medium, and hard.
Table \ref{tab:data_distribution} shows the number and proportion of corpora of different difficulty levels of the \datasetname~training and testing sets.

\textbf{Easy} (number of bounding boxes $\leq 2$ and token length $\leq 16$) contain fewer bounding boxes and shorter texts, resulting in a relatively easy translation task.

\textbf{Hard} (number of bounding boxes $> 5$ or length of tokens $> 25$) contain more bounding boxes or longer texts, which places higher demands on the model's attention mechanism.

\textbf{Medium} (other cases) have a wider spread in terms of the number of bounding boxes and length of tokens, indicating more realistic and complex image translation tasks. With this category, the generalization ability of multimodal translation models can be better tested.

\begin{table*}[t]
\centering

\renewcommand{\arraystretch}{1.1}
\resizebox{\textwidth}{!}{
\begin{tabular}{lcccc|c|ccc|c|ccc|c}
\toprule
\bf Model & \bf Size  &\multicolumn{4}{c}{\bf BLEU} & \multicolumn{4}{c}{\bf chrF++} & \multicolumn{4}{c}{\bf METEOR} \\

\toprule

  &  &\bf Base & \bf Large & \bf Small & \bf Avg  & \bf Base & \bf Large &  \bf Small & \bf Avg   & \bf Base & \bf Large & \bf Small  & \bf Avg \\
\hline
EasyOCR\_NLLB & - & 6.4 & 6.3 & 6.2 & 6.3 & 19.7 & 19.1 & 17.9 & 18.9 & 17.6 & 16.4 & 14.9 & 16.3 \\
\hline
\deepseekvl~ & 7B& 5.9 & 5.7 & 5.5 & 5.7 & 15.5 & 15.1 & 14.8 & 15.1 & 10.9 & 10.5 & 10.2 & 10.5 \\

\llavanext~ & 7B & 8.5 & 8.4 & 8.3 & 8.4 & 20.5 & 20.4 & 20.2 & 20.4 & 13.8 & 13.6 & 13.4 & 13.6 \\

\qwenvl~  & 7B& 14.7 & 14.6 & 13.6 & 14.3 & 29.2 & 29.1 & 28.8 & 29.0 & 21.6 & 21.4 & 21.0 & 21.3 \\

\cogvlm~  &19B & 8.6 & 8.4 & 8.0 & 8.3 & 19.9 & 19.5 & 19.2 & 19.5 & 14.5 & 14.4 & 14.2 & 14.4 \\

\minicpm~ &8B & 11.0 & 9.8 & 9.6 & 10.1 & 22.9 & 21.4 & 20.7 & 21.7 & 17.1 & 16.2 & 15.6 & 16.3\\
\internvl~ &8B & 14.3 & 14.1 & 13.4 & 13.9 & 27.4 & 26.6 & 26.5 & 26.8 & 23.8 & 23.3 & 22.8 & 23.3 \\
\hline
\bf Avg& & \bf 9.9 & \bf 9.6 & \bf 9.2 & \bf 9.6 & \bf 22.1 & \bf 21.6 & \bf 21.1 & \bf 21.6 & \bf 17.0 & \bf 16.6 &  \bf 16.0 & \bf 16.5 \\
\bottomrule
\end{tabular}
}
\caption{Comparison of the results of the image translation. It shows the results of comparison and evaluation of BLEU, chrF++, and METEOR using cascaded and end-to-end IT models on the \datasetname~test set.}
\label{tab:model_comparison}
\end{table*}

\subsection{Diversity}

As shown in Figure~\ref{fig:diversity}, \datasetname~is divided into 28 categories based on image content, ranging from daily objects (e.g., Home \& Kitchenware, Health \& Personal Care) to specialized equipment (e.g., Industrial \& Scientific, Jewelry) and digital goods (e.g., Games \& Apps, Digital Music).
The detailed categories and data can be found in Table \ref{tbl:appendix_data_cate}.
And categories such as "Cell Phones \& Accessories" (162K images) and "Clothing \& Shoes" (86K images) have a significant number of images. 
The products in these categories are frequently used in daily life, which means they are more common. 
This diversity provides a wealth of scenarios for image translation models and enables a comprehensive evaluation of their performance in different image types and text contexts. 
Consequently, training with such diverse data helps in developing models with superior generalization abilities.
The English images dominate the dataset (about 49\%), while Chinese is the second most common language and accounts for about 60\% of the English data. This diversity of language and image data helps to improve the generalization ability of the model in different cultural and linguistic environments.

\begin{figure}[t]
\centering
    \includegraphics[width=0.95\columnwidth]{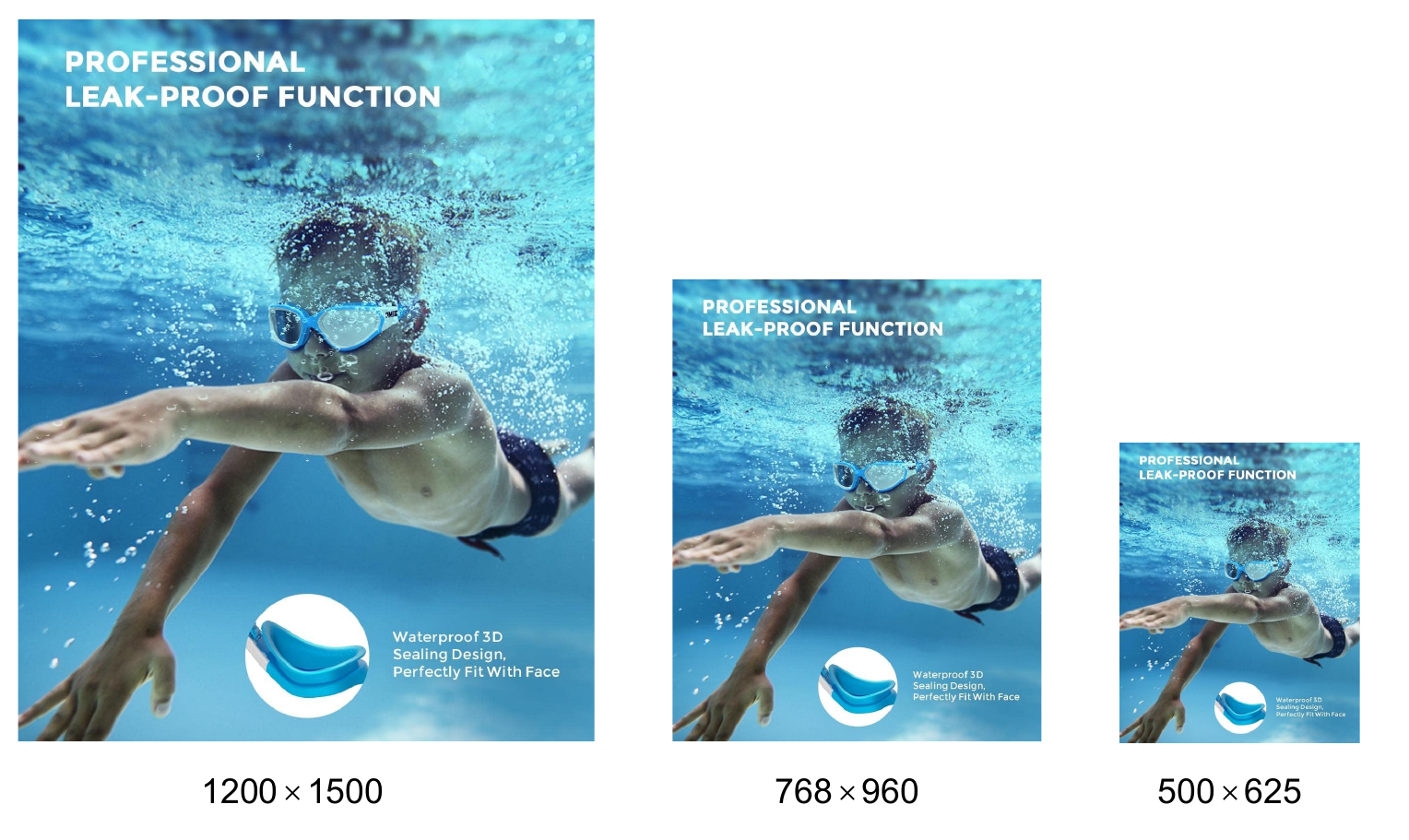}
\caption{Example of images with different resolutions. }
\label{fig:img_quality}
\end{figure}

\subsection{Image resolution}
\label{sec:image_quality}
When translating images to text, the quality and resolution of the images have a direct impact on how the model extracts semantic features from the visual information, which in turn affects translation accuracy.

Unlike previous works, which often only include images with a single resolution (e.g. the ECOIT dataset with an image resolution of only 64x600), \datasetname~selects images with a resolution of more than 1000x1000 pixels (width) when creating the dataset. 
It also provides images in three different sizes: the original size, a "large" version with a width of 768 pixels and a "small" version with a width of 500 pixels ( Figure~\ref{fig:img_quality}).
These different image sizes allow the model to learn visual features under different conditions, which increases its robustness.

\section{Experiments}

In this section, we empirically demonstrate the efficacy of the \datasetname~dataset both as a pretraining dataset as well as an evaluation set for  image-text translate task.

\subsection{Experiment Details}
\paragraph{Setup.}

The operating system which we use is CentOS Linux release 7.5, and the programming language is Python 3.9.12. 
Our experiments are conducted on NVIDIA TESLA A100-40G GPU, the CUDA version is 12.2, and the deep learning framework is torch with version 2.1.0, torchvision with version 0.16.0 and Transformers with 4.44.2.
During the inference stage, we set the following hyperparameters: \textit{temperature} to 0.2.
\paragraph{Models.}

In this study, we perform a comparative analysis of our dataset using both cascaded and end-to-end models for image translation. 
The cascaded model first applies EasyOCR to extract text from images and then translates the extracted text using the NLLB~\cite{nllb2022} model. 
This choice of established components makes our baseline representative of typical cascaded methods and facilitates reproducibility. 
As for the end-to-end model, we compare with mainstream MLLM. 
The detailed introduction is as follows:
\begin{itemize}
\item \textbf{\llavanext}~\cite{liu2024llavanext} is a large-scale multimodal model that can process a variety of data types, including text, images and video. It has been trained on large multimodal datasets and shows strong performance in understanding multiple images and videos. In addition, LLaVA-NeXT has multi-language support, enabling it to understand multilingual text in images, including most European languages, Japanese, Korean, Arabic, Vietnamese and more.
\item \textbf{\qwenvl}~\cite{bai2023qwen} can process images with different resolutions and aspect ratios and supports the understanding of multilingual texts. Qwen2-VL has achieved world-leading performance in several visual comprehension benchmarks and has open-source 2B and 7B scale models.
\item \textbf{\cogvlm}~\cite{hong2024cogvlm2} builds on LLaMA3 and shows exceptional performance on multimodal tasks involving images and text, especially when processing long texts and high-resolution images. It has a visual expert module that focuses on complex visual tasks and achieves deep integration of vision and speech through a sophisticated fusion strategy.
\item \textbf{\internvl}~\cite{chen2024internvl} boosts the model's performance through visual encoder improvements, dynamic strategies for high resolutions and high-quality bilingual data sets. It performs admirably on tasks such as OCR, multimodal assessment, mathematical reasoning and dialog with multiple interlocutors.
\item \textbf{\deepseekvl}~\cite{lu2024deepseek} is an open-source multimodal model designed to improve performance in real-world scenarios. It accepts high-resolution images as input and has general multimodal comprehension capabilities that process logic diagrams, web pages, formula recognition, scientific literature and natural images.

\item \textbf{\minicpm}~\cite{yao2024minicpm} is designed for consumer devices and focuses on providing advanced AI capabilities on resource-constrained devices such as cell phones. It can process various types of data, including text and images, and is capable of describing images, answering text questions with one or more images, writing and debugging code, conducting dialogs with multiple images, conducting dialogs to understand videos, formatting JSON, and performing OCR in high resolution.    
\end{itemize}
In order to make the model output the translated content stably, after several attempts, we use the prompt ``\textit{Translate the text in the image from \{src\_lang\} into \{tgt\_lang\}. Please output the translation directly without any explanation or other words:}'' to obtain the inference result of the model.

\paragraph{Metrics.}
We use the \textit{BLEU} score~\cite{papineni2002bleu} to assess the similarity between predicted translations and reference translations. 
This method calculates n-gram precision and applies a brevity penalty for shorter translations.
We compute the \textit{chrF++} score~\cite{popovic-2017-chrf}, which operates on both character and word-level n-grams. 
It effectively handles morphologically complex languages and is highly sensitive to spelling errors and minor mistakes.
We employ the \textit{METEOR} metric~\cite{banerjee-lavie-2005-meteor} to evaluate translation quality. 
This metric incorporates stemming, synonym matching, and word reordering to improve its correlation with human evaluation.

\subsection{Evaluate IT models with \datasetname}
In this paper, we systematically evaluate the performance of 7 state-of-the-art models on the \datasetname, focusing on the comparison of cascaded models and multimodal end-to-end models on the image translation task. 
Table \ref{tab:model_comparison} shows the performance of all models on the \datasetname.
Among all models, EasyOCR\_NLLB is the only cascaded model, while the others (e.g. \llavanext, \qwenvl) are multimodal end-to-end models.

The Table \ref{tab:model_comparison} shows that \deepseekvl~model perform poorly in IT task, with BLEU, chrF++, and METEOR of 5.7, 15.1, 10.5, well below average.
Although the BLEU, chrF++ and METEOR values of EasyOCR\_NLLB (6.3, 18.9 and 16.3, respectively) are close to the average, they lag behind the end-to-end models.
In contrast, end-to-end models such as \internvl~and \qwenvl~perform better on IT task, with scores of 13.9, 16.8, 23.3 and 14.3, 29.0, 21.3 respectively, far outperforming those of the other models. 
In particular, \qwenvl~achieved the best performance.
The detail comparison of the BLEU of the individual models in multiple languages can be found in Appendix \ref{sec:appendix_translate}.

We attribute this to the ability of the end-to-end models to generate text outputs directly from the image inputs and effectively utilize the correlations between images and text for deeper feature fusion and contextual understanding, resulting in superior performance in the fluency, consistency, and accuracy of the generated text. 
In contrast, cascaded models suffer from error propagation from the OCR phase, which limits their final translation quality.
It is noteworthy that all models show relatively low evaluation results compared to their performance on other datasets. 
This highlights the realism of the  dataset, which includes a wider range of challenges, including text with different fonts, colors and backgrounds, as well as instances of text occlusion and blurring. 
Consequently, the \datasetname~provides a more realistic and sophisticated benchmark for assessing the generalization ability of models.

\begin{table}[t]
  \centering
  \renewcommand{\arraystretch}{1.1}
  
  \resizebox{0.95\linewidth}{!}{
    \begin{tabular}{lcccccc}
    \toprule
    &\textbf{Easy} & \textbf{Medium} & \textbf{Hard} \\
    \midrule        
    EasyOCR\_NLLB & 19.2 & 18.9 & 10.0 \\
    \hline
    \deepseekvl  & 10.9 & 10.6 & 10.0 \\
    \llavanext & 15.6 & 14.3 & 10.8 \\
    \qwenvl  & 23.3 & 21.5 & 18.8 \\
    \cogvlm & 15.6 & 14.6 & 12.5 \\
    \minicpm & 17.7 & 15.0 & 15.8 \\
    \internvl~& 25.9 & 23.2 & 19.9 \\
    \hline
    \bf Avg &\bf 18.3 &\bf 16.9 &\bf 14.0 \\        
    \bottomrule
  \end{tabular}
  }
 \caption{METEOR with different levels of difficulty.}
\label{tab:difficulty_comparison} 
\end{table}

\begin{figure*}[t]
    \centering
    \includegraphics[width=\linewidth]{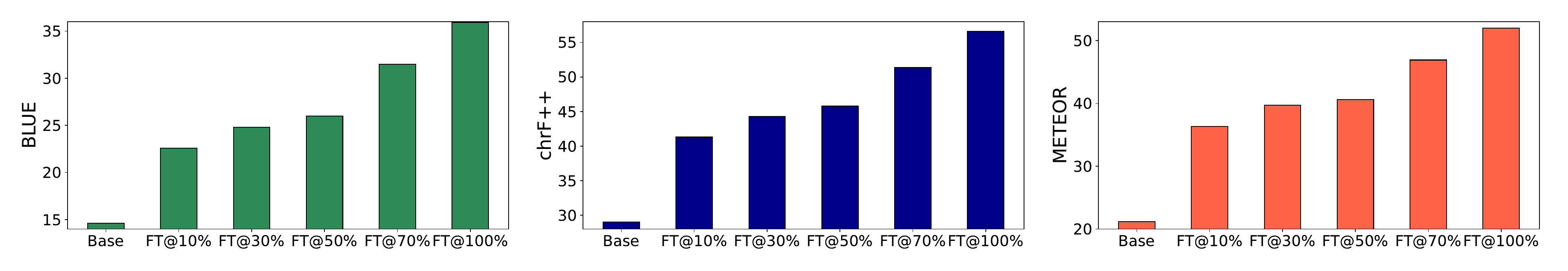}
    \caption{The performance comparison of the 5 model in the metrics: BLEU, chrF++, and METEOR. The 5 models are the base version and the version fine-tuned with 10\%, 30\%, 50\%, 70\%, and 100\% \datasetname~train set.}
    \label{fig:sft}
\end{figure*}

\subsection{Effect of Resolution and Difficulty}

To further investigate the properties of the \datasetname, we conducted finer analyzes at different image resolutions and task difficulties.

Table \ref{tab:model_comparison} illustrates the performance differences of the models in the image resolution tasks. 
A comparison of the average score of all models in different sizes shows that almost all models perform better with larger images. 
The BLEU for the base, large and small resolutions are 9.9, 9.6 and 9.2 respectively, while the chrF++ and METEOR are 22.1, 21.6, 21.1 and 17.0, 16,6, 16.0 respectively.
This emphasizes the importance of image resolution for the model's understanding of the image content. High-resolution images provide more detailed information that helps the model to recognize and translate the text in the image more accurately.

Furthermore, \datasetname~contains tasks with three levels of difficulty: \textit{Easy}, \textit{Medium} and \textit{Difficult} (see \S Section \ref{sec:diff_levels}). 
Table \ref{tab:difficulty_comparison} illustrates the differences in model performance for the different tasks. 
When analyzing the performance of the models on the different difficulty levels, we observe a clear trend that the performance decreases with increasing difficulty. 
The average METEOR drops from 18.3 to 14.0, and this is particularly evident for \qwenvl, whose score decreases from 23.3 to 18.8. 

This shows that the carefully designed difficulty distribution in the \datasetname~can effectively differentiate the performance of the models and provide the models with different difficulty levels.

As for the evaluation metrics, given the abundance of short sentences in \datasetname, we recommend focusing on chrF++ and METEOR in addition to BLEU, as they are more suitable for evaluating the quality of short sentence translations.

Overall, our experimental results show that \datasetname~is a comprehensive, high quality and realistic dataset for image to text translation. \datasetname~represents a challenging benchmark for image translation research and serves as a valuable resource to advance the development of robust and versatile multimodal models.

\subsection{Fine-tune with \datasetname}

To evaluate the effectiveness of our proposed \datasetname~for training image-to-text translation models, we perform fine-tuning experiments using the \qwenvl~model as the baseline. 
During training, we use the following important hyperparameter settings: \texttt{per\_device\_train\_batch\_size}, \texttt{gradient\_accumulation\_steps}, \texttt{learning\_rate}, \texttt{num\_train\_epochs}, \texttt{lr\_scheduler\_type} and \texttt{warmup\_ratio} are set to 4, 32, 1.0e-4, 1, cosine and 0.1, respectively. 
Additionally, we utilize mixed precision training to accelerate the training process and reduce memory consumption.

We fine-tune the model with different subsets of the \datasetname~to investigate the effects of data size on model performance.
Specifically, we take 10\%, 30\%, 50\%, 70\%, and 100\% of the training set from the \datasetname~ and compare their performance to analyze the effects of data size.

As shown in Figure \ref{fig:sft}, the results demonstrate that increasing the size of the training data leads to significant improvements in all three metrics. For example, the BLEU increases from 14.6 for the base model (base@10\%) to 35.9. Similarly, the chrF++ score increases continuously from 29.0 to 56.6 and the METEOR score improves from 21.2 to 52.0.
It strongly emphasize the crucial role of data size in improving the performance of the image-to-text translation model: with richer training data, the model can capture finer correlations between images and text, improving translation accuracy and fluency.
It is noteworthy that even with only 10\% of the training data, the performance of the model is still significantly better than that of the base model, indicating the high quality of the \datasetname. This result emphasizes that our dataset can effectively improve the performance of the model even with relatively small amounts of data.

\subsection{Comparison with Existing IT Datasets}

\begin{table}[t]
\centering
  \renewcommand{\arraystretch}{1.1}
  
    \resizebox{0.95\linewidth}{!}{
    \begin{tabular}{lcccc}
    \toprule
    Dataset & DE-EN & EN-DE & EN-FR & FR-EN \\
    \midrule
    IIMT \cite{lan-etal-2024-translatotron} & 14.5 & 15.0 & 16.4 & 21.5 \\
    \bf MIT-10M (ours) & \bf 16.4 &\bf 16.2 &\bf 19.3 &\bf 25.3 \\
    \bottomrule
    \end{tabular}
}
\caption{BLEU of fine-tuning on subsets of IIMT (170K) and \datasetname~(150K).}
\label{tab:finetuning_comparison}
\end{table}

To further demonstrate the advantages of \datasetname~over existing real-world datasets, we perform fine-tuning experiments with comparable subsets.
We compare the fine-tuning performance of \datasetname~with the IIMT dataset. We select subsets of both datasets to control the data size. 
We use 150K image-text pairs from \datasetname~and the slightly larger 170K pairs from IIMT. 
This choice ensures that any performance differences are not simply attributable to the volume of training data, but rather to inherent dataset qualities.  
We focus on four language pairs: DE-EN, EN-DE, EN-FR, and FR-EN. 
The BLEU after fine-tuning can be found in Table \ref{tab:finetuning_comparison}.
Even with less data, \datasetname~consistently outperforms IIMT in all four language pairs. 
The performance gains are particularly noticeable in the FR-EN pair, where \datasetname~shows a nearly 4 BLEU improvement. 
The results strongly support that \datasetname~is a valuable resource for multilingual image translation, enabling the training of more robust and generalizable models.  
This comparative analysis provides compelling evidence that the quality and diversity of \datasetname~contribute significantly to improved fine-tuning results, even when controlling for data size.

\section{Conclusion}
In this paper, we propose \datasetname, a large-scale, high-quality dataset designed for multilingual image translation.
\datasetname~ contains over 10M image-text pairs in 14 languages and 840K high-resolution real-world images.
We described the dataset creation process in detail and conducted a comprehensive analysis of the dataset across multiple dimensions.
The results of experiments with different end-to-end IT models and evaluation metrics show that \datasetname~significantly improves multilingual translation performance and has strong generalization capabilities, especially for complex tasks.
In the future, we will focus on further improving translation accuracy, extending support for more languages and processing even more diverse and complex images.

\section*{Acknowledgments}
The present research was supported by the National Key Research and Development Program of China (No. 2024YFB3309702) and the National Natural Science Foundation of China Youth Fund (Grant No.62306210).

\section*{Limitations}
In this work, we introduce \datasetname, a novel, large-scale multilingual image translation parallel corpus that significantly advances research in the field of cross-lingual image translation.
However, our approach is not without limitations.
First, it is a major challenge to achieve a balanced representation of language and domain within the dataset.
Furthermore, despite careful annotation and translation, the inherent complexity of multilingual data may lead to inaccuracies that could affect the reliability of the dataset.
Furthermore, while the dataset has been extensively cleaned and filtered to address ethical concerns, including the removal of privacy and sensitive content issues, unforeseen possibilities remain.
We acknowledge these limitations transparently to promote ethical research and encourage the community to make improvements.

\bibliography{main}

\appendix

\section{Prompt}
\subsection{OCR Prompt}
\label{sec:appendix_ocr_prompt}

To enhance the precision of text identification within images, we employ the GPT-4o model for Optical Character Recognition (OCR) across the dataset. 
We meticulously construct a prompt for this task, ensuring that the textual data extracted is both accurate and reliable for subsequent analysis and translation processes. 
It is important to note that we instruct the GPT-4o model to output both English and Chinese content. This bilingual output capability is used to identify and translate the text, which is essential for the subsequent calibration phase.
\begin{lstlisting}
Please perform text recognition on an image and translate the recognized text into other language.The output should be in json format\:
'lang' refers to the original language of the recognized text, such as 'es', 'de', etc.
'text' is the text recognized from the image. If there are multiple lines in the original text, please output them line by line.
'en' is the result translated into English, corresponding line by line with the original text.
'zh' is the result translated into Simplified Chinese. 
\end{lstlisting}

\subsection{Translation Prompt}
\label{sec:appendix_trans_prompt}
We use the GPT-4 model to translate the extracted text into 13 target languages.
To improve the quality of the translations, we use carefully crafted prompts. We also call Google Translate with the same prompts for cross-validation. 
The translations are further refined by filtering out those with ambiguous meanings based on semantic similarity scores to ensure the accuracy and reliability of the translations.
\begin{figure}[t]
\begin{lstlisting}
Translate the following sentence into multiple languages, keeping the original paragraph structure and translating line by line. 
Sentence:```text```
Using this JSON schema:
Lang = {"hi": str, "tr": str, "zh": str, "en": str, "fr": str, "de": str, "ja": str, "it": str,"ko": str, "th": str, "ru": str, "pt": str,"en": str,"ar": str}
Return a `dict[str, Lang]`
'en' is English.
'zh' is Chinese (Simplified).
'de' is German.
'fr' is French.
'ja' is Japanese.
'it' is Italian.
'ko' is Korean.
'th' is Thai.
'ru' is Russian.
'pt' is Portuguese.
'es' is Spanish.
'hi' is Hindi.
'tr' is Turkish.
'ar' is Arabic.
\end{lstlisting}
\end{figure}

\section{Dataset Examples}
\label{sec:appendix_examples}
\begin{figure*}[t]
    \centering
    \includegraphics[width=\textwidth]{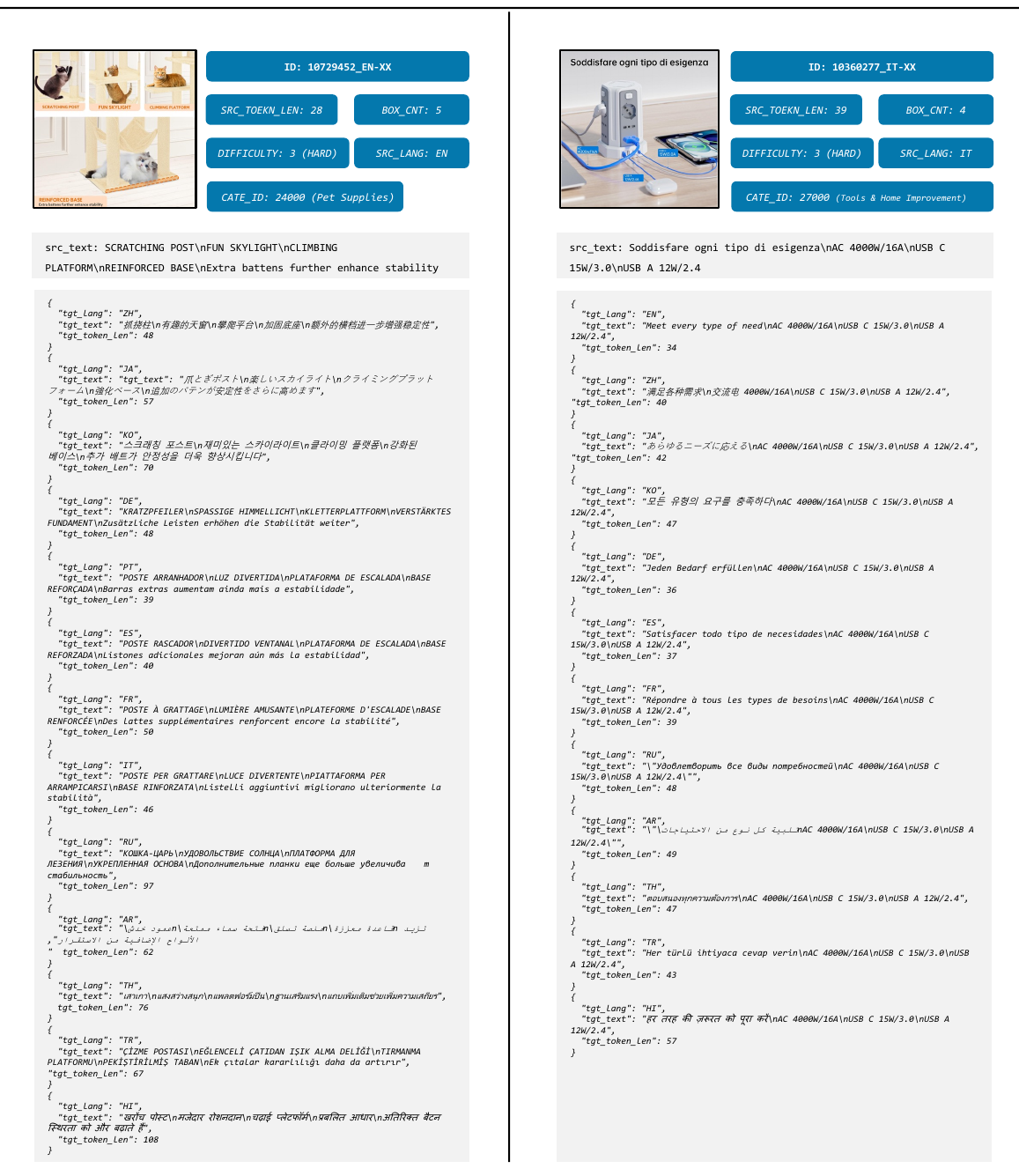}
    \caption{The detail field description for the \datasetname~dataset.}
    \label{fig:full_demo}
\end{figure*}

Figure \ref{fig:full_demo} shows the detail fields in the dataset, including 6 labels and 14 languages text.

\section{Experiments}

\label{sec:appendix_translate}
\begin{figure*}[t]
    \centering
    
    \begin{subfigure}{0.24\textwidth}
        \includegraphics[width=\textwidth]{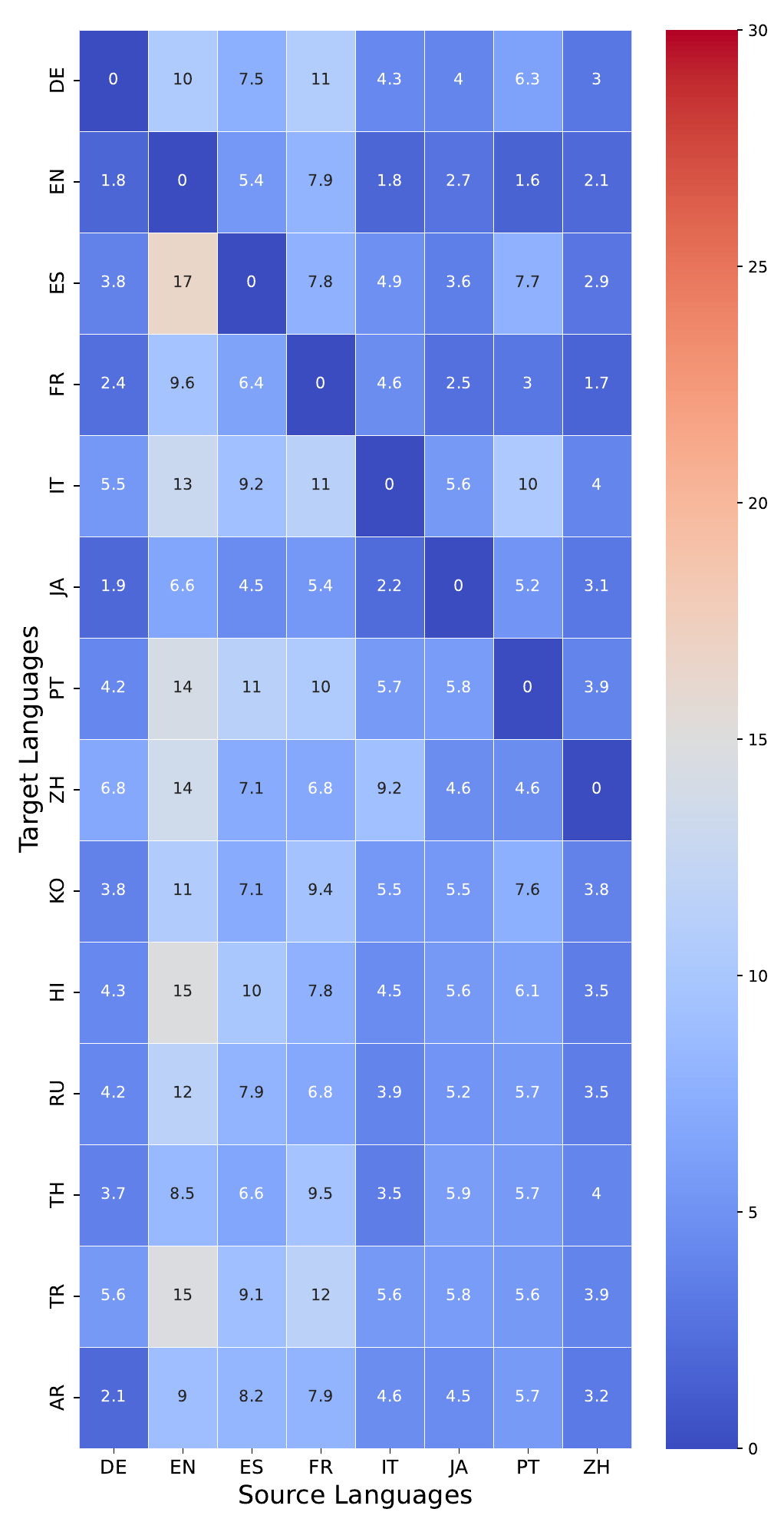}
        \caption{EasyOCR\_NLLB}
    \end{subfigure}
   \begin{subfigure}{0.24\textwidth}
        \includegraphics[width=\textwidth]{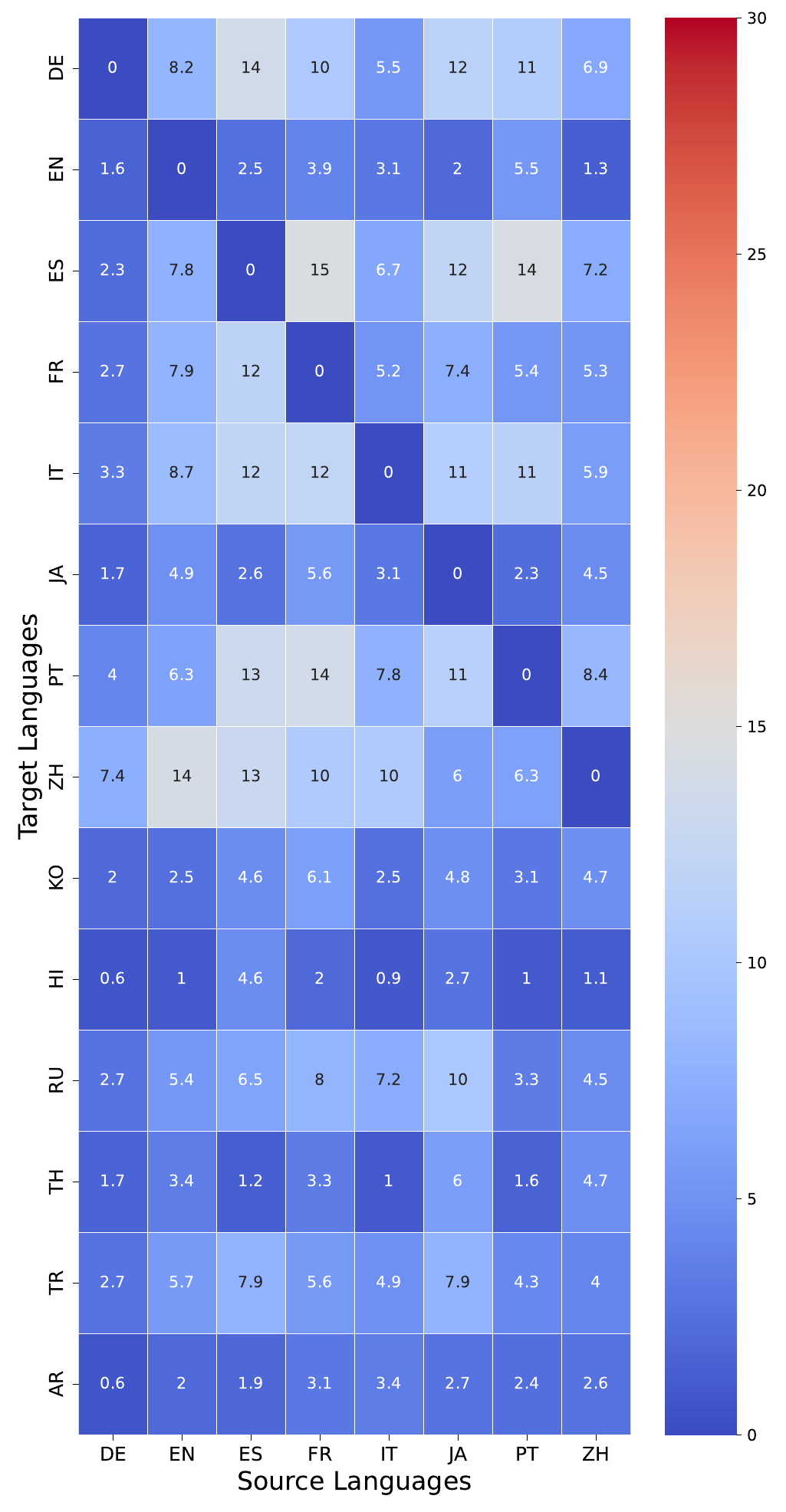}
        \caption{\deepseekvl}
    \end{subfigure}
    \begin{subfigure}{0.24\textwidth}
        \includegraphics[width=\textwidth]{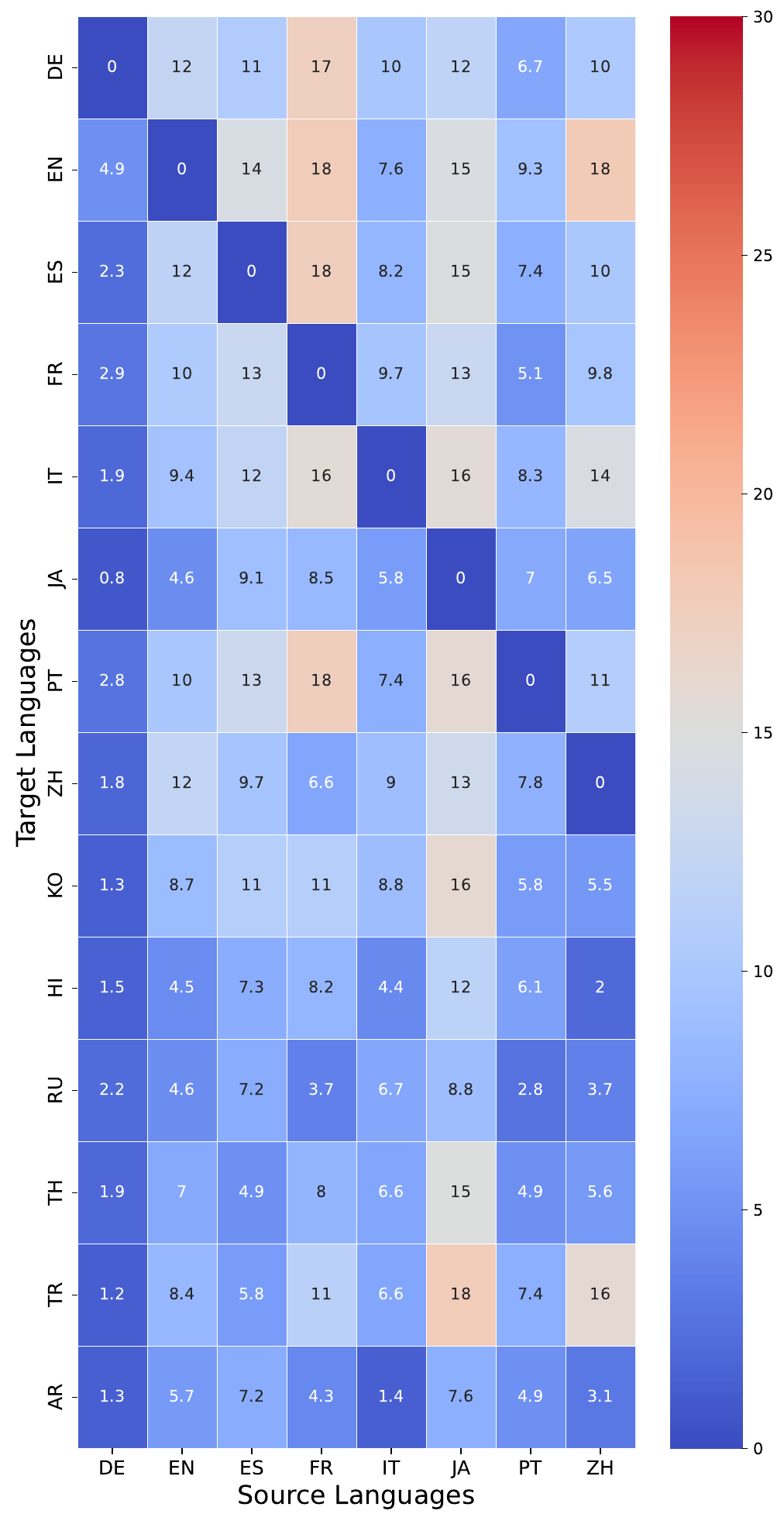}
        \caption{\llavanext}
    \end{subfigure}
   \begin{subfigure}{0.23\textwidth}
        \includegraphics[width=\textwidth]{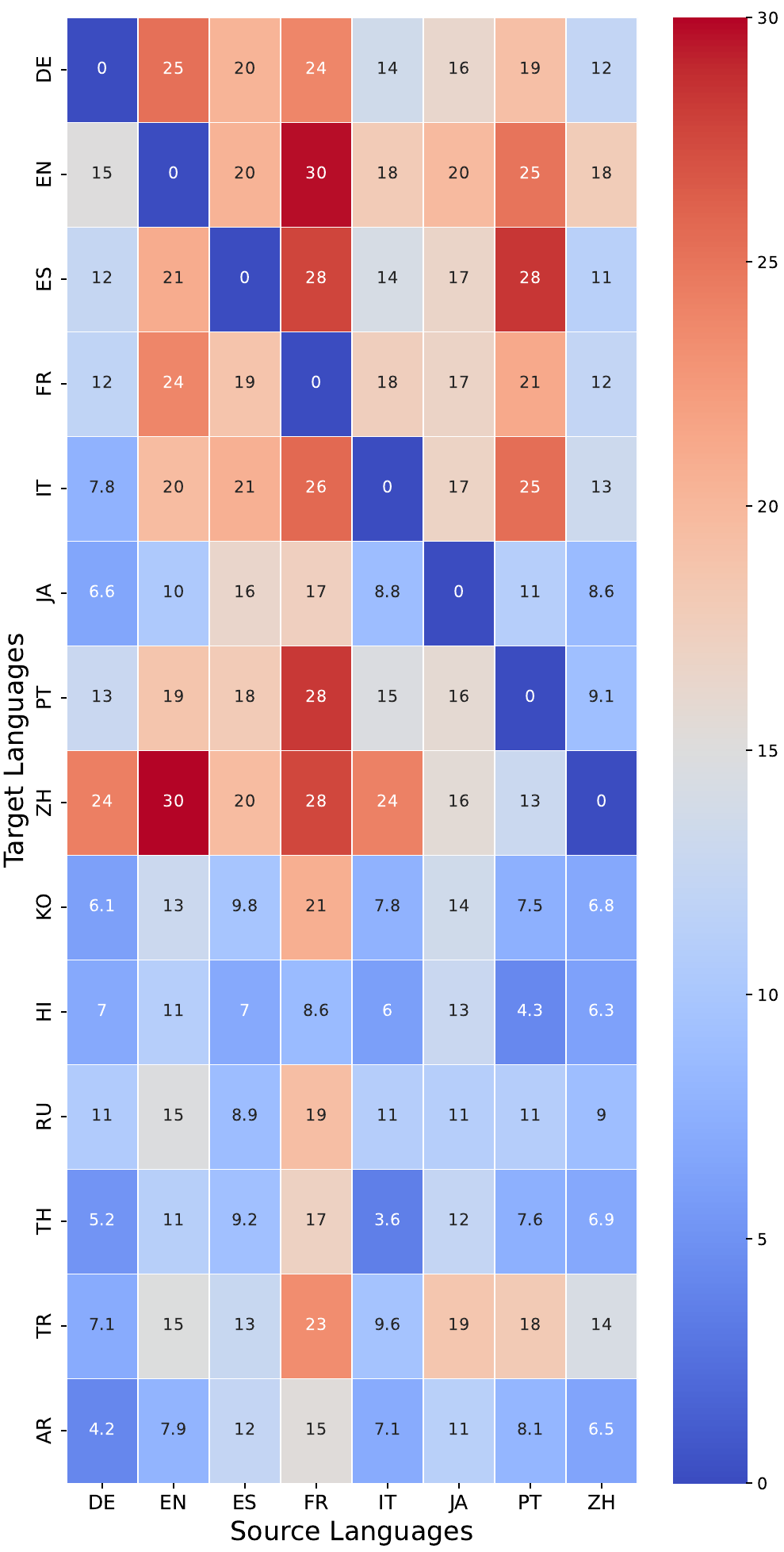}
        \caption{\qwenvl}
    \end{subfigure}
    \begin{subfigure}{0.24\textwidth}
        \includegraphics[width=\textwidth]{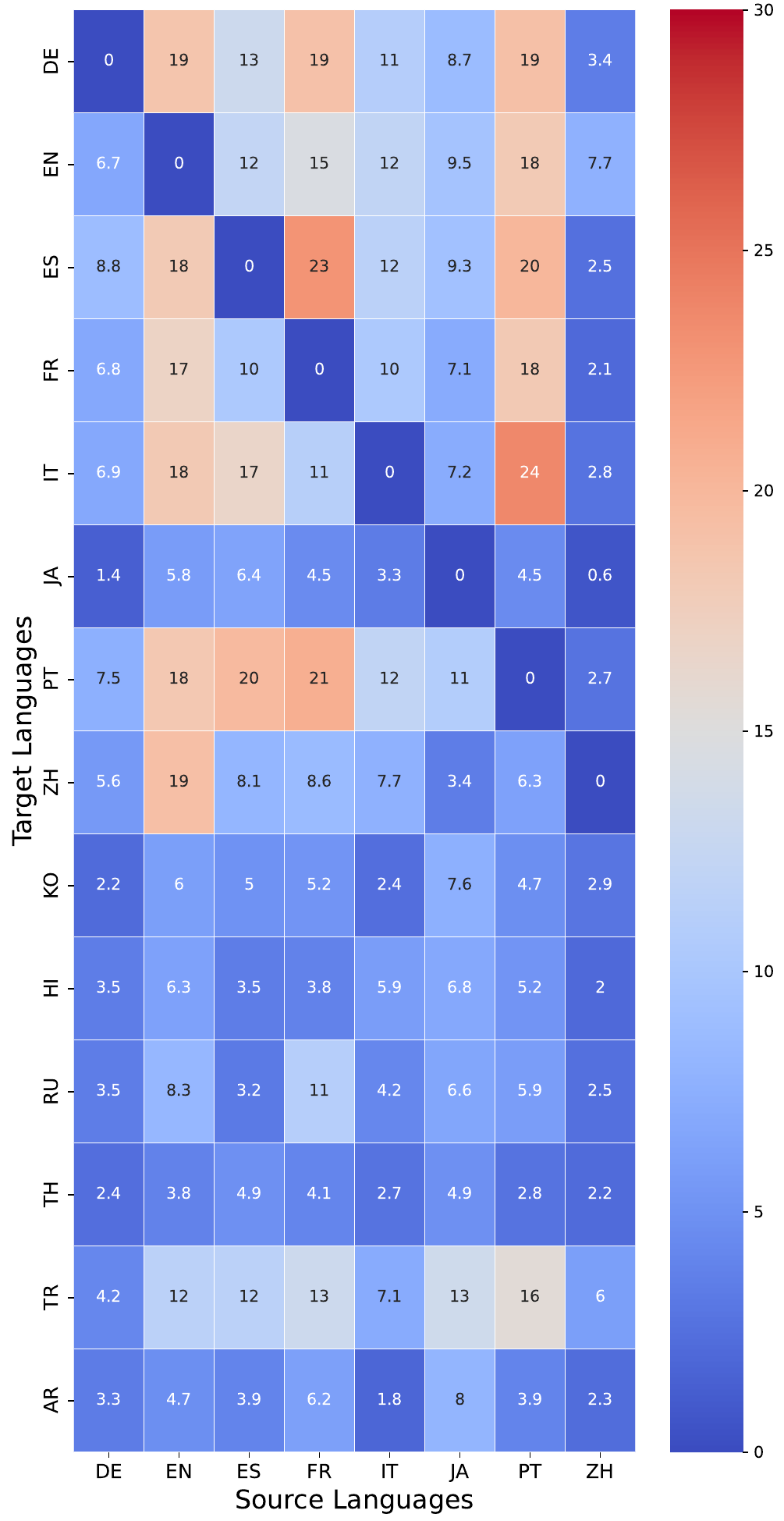}
        \caption{\cogvlm}
    \end{subfigure}
   \begin{subfigure}{0.24\textwidth}
        \includegraphics[width=\textwidth]{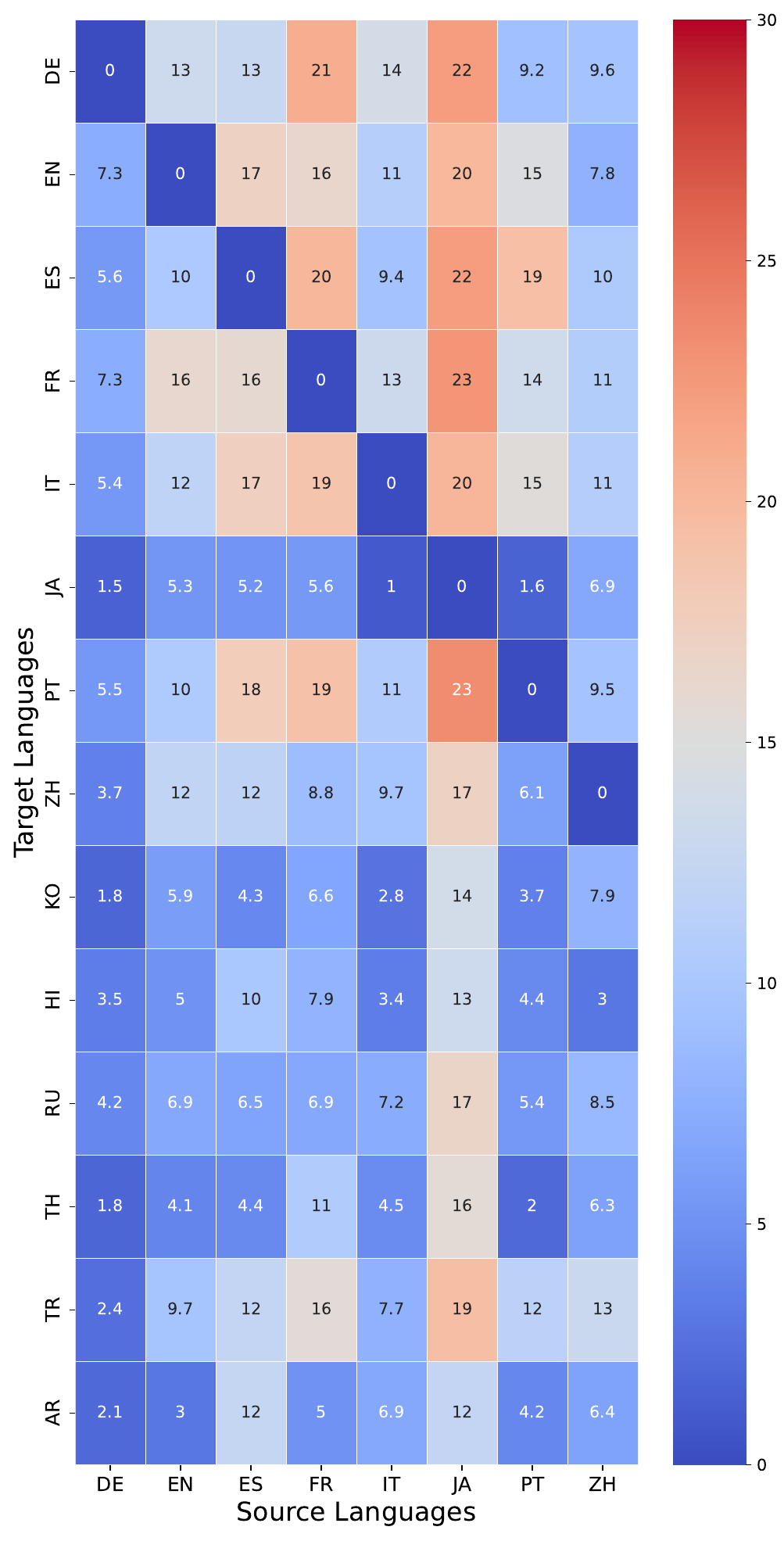}
        \caption{\minicpm}
    \end{subfigure}
    \begin{subfigure}{0.23\textwidth}
        \includegraphics[width=\textwidth]{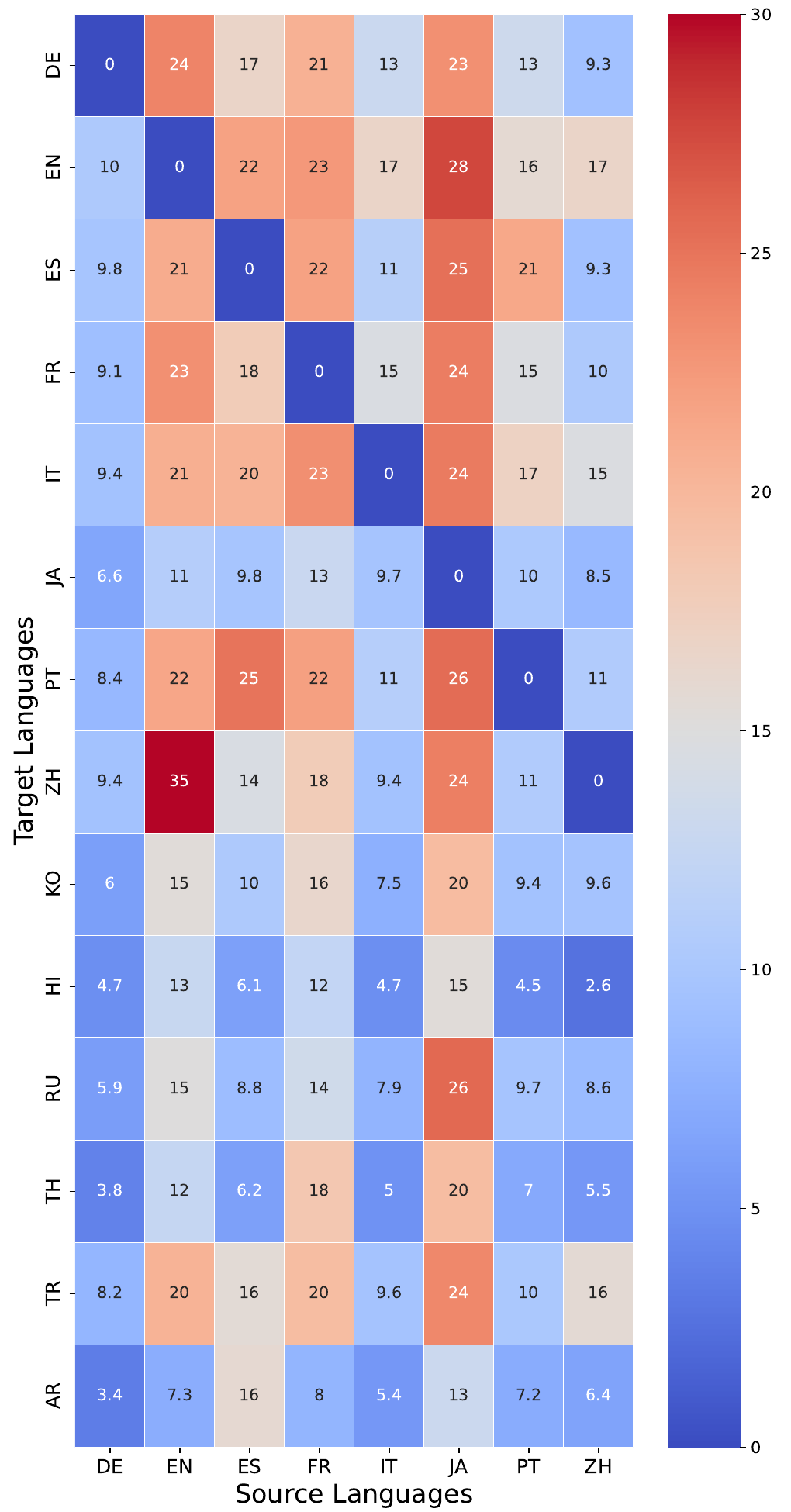}
        \caption{\internvl}
    \end{subfigure}
    
    \caption{The heatmap of various models in image translation tasks.}
    \label{fig:heatmap}
\end{figure*}

Figure \ref{fig:heatmap} show the detail comparison of the BLEU of each model in multiple language pairs.

\end{document}